\documentclass[11pt,a4paper]{article}
\usepackage[hyperref]{acl2021}
\usepackage{times}
\usepackage{latexsym}

\usepackage{hyperref}
\usepackage{graphicx}
\usepackage{bm}
\usepackage{amsmath}
\usepackage{amsfonts}
\usepackage{color}
\usepackage{url}
\usepackage{enumitem}
\setlist{nosep}
\newcommand{\tabincell}[2]{\begin{tabular}{@{}#1@{}}#2\end{tabular}}
\usepackage{multirow}
\usepackage{multicol}
\usepackage{url}
\usepackage{subfig}
\usepackage{soul}
\usepackage{pifont}
\setenumerate{leftmargin=*}

\usepackage[autostyle]{csquotes}

\makeatletter
\def\thickhline{%
  \noalign{\ifnum0=`}\fi\hrule \@height \thickarrayrulewidth \futurelet
   \reserved@a\@xthickhline}
\def\@xthickhline{\ifx\reserved@a\thickhline
               \vskip\doublerulesep
               \vskip-\thickarrayrulewidth
             \fi
      \ifnum0=`{\fi}}
\makeatother

\newlength{\thickarrayrulewidth}
\usepackage{microtype}

\aclfinalcopy % Uncomment this line for the final submission
\def\aclpaperid{1535} %  Enter the acl Paper ID here

\newcommand{\gab}[1]{\textcolor{black}{{#1}}}
\newcommand{\yh}[1]{\textcolor{black}{{#1}}}

\newcommand{\cmrdy}[1]{\textcolor{black}{{#1}}}

\title{Topic-Driven and Knowledge-Aware Transformer for Dialogue Emotion Detection}
\author{
 Lixing Zhu\textsuperscript{\dag}, Gabriele Pergola\textsuperscript{\dag}, Lin Gui\textsuperscript{\dag}, Deyu Zhou\textsuperscript{\S}, Yulan He\textsuperscript{\dag} \\
 \textsuperscript{\dag}Department of Computer Science, University of Warwick, UK\\
 \textsuperscript{\S}School of Computer Science and Engineering, Key Laboratory of Computer Network\\
 and Information Integration, Ministry of Education, Southeast University, China\\
 {\tt \{lixing.zhu,gabriele.pergola,lin.gui,yulan.he\}@warwick.ac.uk} \\
 {\tt d.zhou@seu.edu.cn}
}

\date{}

\begin{document}
\maketitle
\begin{abstract}
Emotion detection in dialogues is challenging as it often requires the identification of thematic topics underlying a conversation, the relevant commonsense knowledge, and the intricate transition patterns between the affective states. In this paper, we propose a Topic-Driven Knowledge-Aware Transformer to handle the challenges above. We firstly design a topic-augmented language model (LM) with an additional layer specialized for topic detection. The topic-augmented LM is then combined with commonsense statements derived from a knowledge base based on the dialogue contextual information. Finally, a transformer-based encoder-decoder architecture fuses the topical and commonsense information, and performs the emotion label sequence prediction. The model has been experimented on four datasets in dialogue emotion detection, demonstrating its superiority empirically over the existing state-of-the-art approaches. Quantitative and qualitative results show that the model can discover topics which help in distinguishing emotion categories.
\end{abstract}

\section{Introduction}
The abundance in dialogues extracted from online conversations and TV series %have provided abundant corpora for, facilitate
provides unprecedented opportunity to train models for automatic emotion detection, which are important for the development of empathetic conversational agents or chat bots for psychotherapy~\cite{hsuku2018socialnlp,jiao2019higru,zhang2019modeling,cao2019observing}. However, it is challenging to capture the contextual semantics of personal experience described in one's utterance. For example, the emotion of the sentence ``\emph{I just passed the exam}'' can be either \emph{happy} or \emph{sad} depending on the expectation of the subject. There are strands of works utilizing the dialogue context to enhance the utterance representation~\cite{jiao2019higru,zhang2019modeling,majumder2019dialoguernn}, where influences from historical utterances were handled by recurrent units, and attention signals were further introduced to intensify the positional order of the utterances.

% \gab{The recently increased engagement of automatic conversational agents in everyday life has inevitably called for interactions that are oriented at satisfying user's requests on particular topics while being simultaneously empathetic and enjoyable; a characteristic that becomes even more relevant in delicate and multifaceted situations where chat bots are, for example, employed for psychotherapy ~\cite{hsuku2018socialnlp,cao2019observing}.
% The widely availability of dialogues extracted from online conversations and TV series has recently made possible promising research employing neural models as conversational agents \cite{jiao2019higru,zhang2019modeling, majumder2019dialoguernn}. They are mainly focused on modelling} the dialogue context to enhance utterance representations via \gab{recurrent mechanism to model the dialog development} and attention mechanisms to \gab{weights the relevant parts the mostly influence the agent interactions} ~\cite{jiao2019higru,zhang2019modeling,majumder2019dialoguernn}.

% % As the utterance shown in an example ``'', the
% \begin{figure}[tp]
% \centering
% \includegraphics[width=0.48\textwidth]{./figs/fig0}
% \caption{Utterances around particular topics carry specific emotions in the DailyDialog dataset. \cmrdy{Utterances carrying \emph{positive} (smiling face) or \emph{negative} (crying face) emotions are highlighted in colour. Other utterances are labeled as `\emph{Neutral}'.}}
% \label{fig:0}
% \end{figure}

% As the utterance shown in an example ``'', the
\begin{figure}[tp]
\centering
\includegraphics[width=0.48\textwidth]{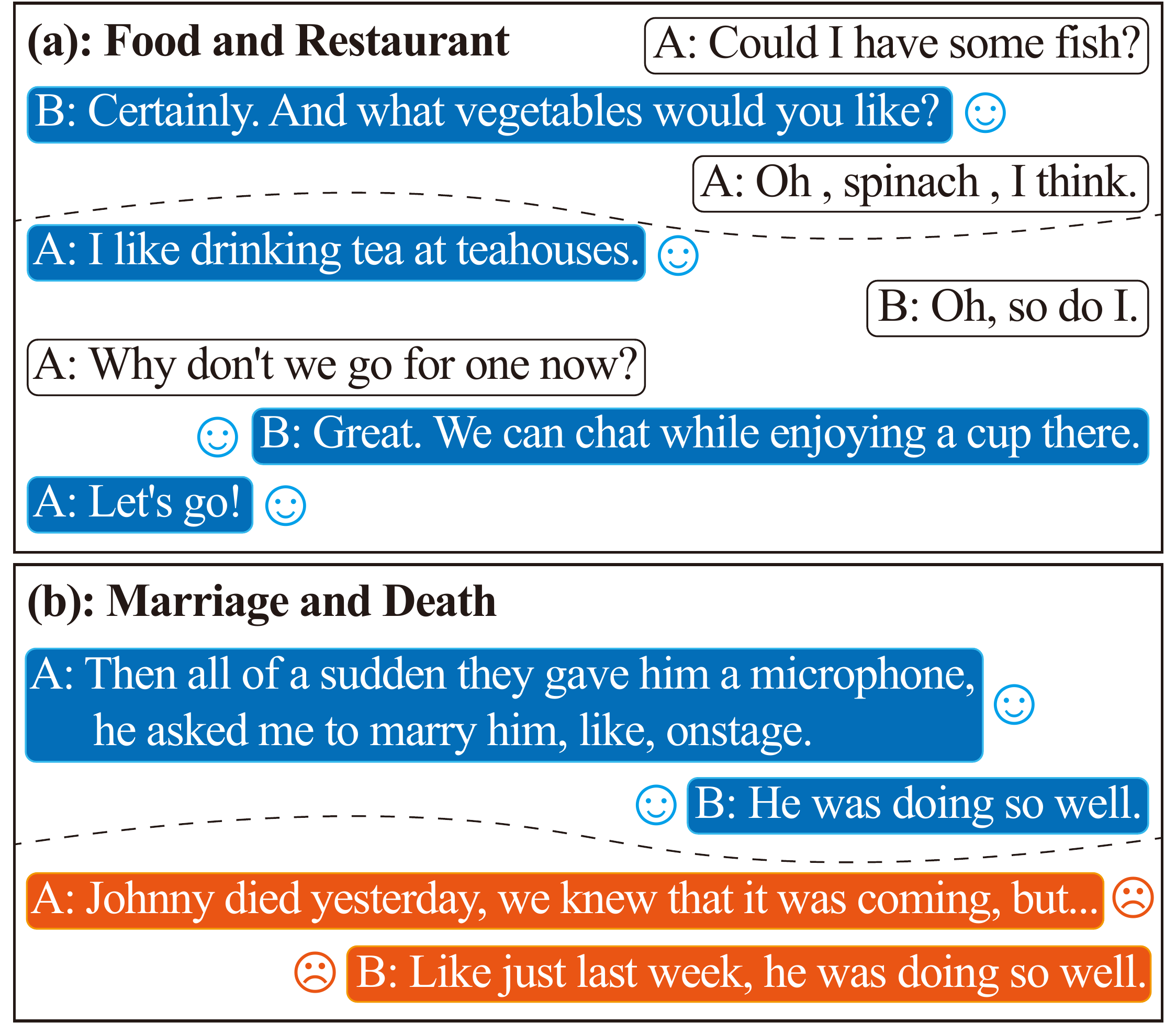}
\caption{Utterances around particular topics carry specific emotions. \cmrdy{Utterances carrying \emph{positive} (smiling face) or \emph{negative} (crying face) emotions are highlighted in colour. Other utterances are labeled as `\emph{Neutral}'.} In (a), utterances discussing food and restaurant are more likely carrying positive sentiment. In (b), the similar utterance, `\emph{He was doing so well}', expressed different emotions depending on its associated topic.}
\label{fig:0}
\end{figure}
% in the DailyDialog dataset

\gab{Despite the progress made by the aforementioned methods, detecting emotions in dialogues is however still a challenging task due to the way emotions are expressed and how the meanings of utterances vary based on the particular topic discussed, as well as the implicit knowledge shared between participants.}
% \gab{For example, the utterance ``\emph{I just passed the exam}'' might convey different emotions (i.e. \emph{happy} or \emph{sad}) depending on the pre-existing expectation on the subject.}  % \gab{instead}
Figure~\ref{fig:0} gives an example of how topics and background knowledge could impact the mood of interlocutors. Normally, dialogues around specific topics carry certain language patterns~\cite{serban2017hierarchical}, \gab{affecting not only the utterance's meaning, but also the particular emotions conveyed by specific expressions.}
% The same correlation also exists for emotion labels.
Existing dialogue emotion detection methods did not \gab{put emphasis on} model\gab{ling} these holistic properties of dialogues (i.e., conversational topics and tones).
\gab{Consequently, they were fundamentally limited in capturing the affective states of interlocutors related to the particular themes discussed.}
% Consequently, they are unable to capture these intrinsic properties and thus deficient in discovering the affect states.
\gab{Besides, emotion and topic detection heavily relies on leveraging underlying commonsense knowledge shared between interlocutors. Although there have been attempts in incorporating it, such as the COSMIC \cite{ghosal2020cosmic}, existing approaches do not perform fine-grained extraction of relevant information based on both the topics and the emotions involved.}
% In addition, emotion detection requires the incorporation of commonsense knowledge. Current approaches such as \citet{ghosal2020cosmic}'s COSMIC lacks effective methods to incorporate the external knowledge. It is worthy of delving into the pattern between knowledge selection and emotion.

% Add a footnote describing the percentage of restaurant.
% Add source code.
% Change IEMOCAP graph
% Average Pooling and w/o attention place them to the text

% which is indispensable for better understanding of hidden semantics.
% affect states % behind one's narratives. % semantic properties

Recently, the Transformer architecture~\cite{vaswani2017attention} has empowered language models to transfer large quantities of data to low-resource domains, making it viable to discover topics in conversational texts. In this paper, we propose to add an extra layer to the pre-trained language model to model the latent topics, which is learned by fine-tuning on dialogue datasets to alleviate the data sparsity problem. Inspired by the success of Transformers, we use the Transformer Encoder-Decoder structure to perform the Seq2Seq prediction in which an emotion label sequence is predicted given an utterance sequence (i.e., each utterance is assigned with an emotion label). We posit that the dialogue emotion of the current utterance depends on the historical dialogue context and the predicted emotion label sequence for the past utterances. %utterances the model has seen so far and the previous utterance emotions.
We leverage the attention mechanism and the gating mechanism to incorporate commonsense knowledge retrieved by different approaches. Code and trained models are released to facilitate further research\footnote{\url{http://github.com/something678/TodKat}.}.
% \footnote{Code and trained models are available at \url{http://github.com/somethingx02/topical\_wordvec\_models}.}.
% released in supplementary materials to facilitate further research.
To sum up, our contributions are:

\begin{itemize}
\item We are the first to \yh{propose a topic-driven approach} %apply topic modelling
for dialogue emotion detection. We propose to alleviate the low-resource setting by topic-driven fine-tuning using pre-trained language models.
% with latent variables added to discover topics that facilitate emotion detection on dialogue datasets.
\item We utilize a pointer network and an additive attention to integrate commonsense knowledge from multiple sources and dimensions.
\item We develop a Transformer Encoder-Decoder structure as a replacement of the commonly-used recurrent attention neural networks for dialogue emotion detection.
\end{itemize}

% Quick start:
%
% Additionally, w by various means develop, provide, utilize, exploit topic-driven fine-tuning
% to address the  sequence-to-sequence prediction problem
% We provide a latent variable model that fine tunes language models
% propose to leverage
% sequence-to-sequence dialogue emotion
% the commonsense knowledge, we
% At the same time, knowledge are comprehensive supplement interpretation comprehensive
% . however. Normally, dialogues around specific topics carry specific language patterns \cite{serban2017hierarchical}. The same correlation also exists in emotion labels. Different topics indicate different moods. Dialogues are guided or dominated by specific topics

\section{Related Work}
\paragraph{Dialogue Emotion Detection} \newcite{majumder2019dialoguernn} recognized the importance of dialogue context in dialogue emotion detection. They used a Gated Recurrent Unit (GRU) to capture the global context which is updated by the speaker ad-hoc GRUs. At the same time, \newcite{jiao2019higru} presented a hierarchical neural network model that comprises two GRUs for the modelling of tokens and utterances respectively. \newcite{zhang2019modeling} explicitly modelled the emotional dependencies on context and speakers using a Graph Convolutional Network (GCN). Meanwhile, \newcite{ghosal2019dialoguegcn} extended the prior work~\cite{majumder2019dialoguernn} by taking into account the intra-speaker dependency and relative position of the target and context within dialogues. %, which are represented as a directed graph.
Memory networks have been explored in~\cite{jiao2020real} to allow bidirectional influence between utterances. A similar idea has been explored by~\newcite{li2020bieru}. While the majority of works have been focusing on textual conversations, \newcite{zhong2019knowledge} enriched utterances with concept representations extracted from the ConceptNet~\cite{speer2017conceptnet}. %, a supplement for commonsense knowledge, and showed efficacy. Similar to~\cite{zhong2019knowledge},
\newcite{ghosal2020cosmic} developed COSMIC which exploited \textsc{Atomic}~\cite{sap2019atomic} for the acquisition of commonsense knowledge. Different from existing approaches, we propose a topic-driven and knowledge-aware model built on a Transformer Encoder-Decoder structure for dialogue emotion detection. %\cmrdy{The proposed Seq2Seq structure is identical to KET~\cite{zhong2019knowledge} as they both predict emotions by taking into account the past utterances and emotions.}
%n that emotions are predicted taking into account both the historical utterances and emotions and that a decoder is employed to handle these contexts.}

% where in dialogue emotion. take into account.  Other work that including .

% Meanwhile, DRcrt, COSMIC blablabla

\paragraph{Latent Variable Models for Dialogue Context Modelling} Latent variable models, normally described in their neural variational inference form named \yh{Variational Autoencoder} (VAE)~\cite{kingma2014auto}, \cmrdy{has been studied extensively to learn thematic representations of individual documents~\cite{miao2016neural,srivastava2017autoencoding,NEURIPS2020_9f1d5659}. They have been successfully employed for dialogue generation to model thematic characteristics over dynamically evolving conversations.
%while retaining a level of flexibility between conversations.
This line of work, which inlcudes approaches based on hierarchical recurrent VAEs~\cite{serban2017hierarchical,parketal2018hierarchical,zengetal2019dirichlet} and conditional VAEs~\cite{sohn2015learning,shen2018improving,gao2019discrete}, encodes each utterance with historical latent codes and autoregressively reconstructs the input sequence.}

% This line of work includes hierarchical recurrent VAE methods~\cite{serban2017hierarchical,parketal2018hierarchical,zengetal2019dirichlet} and conditional VAE methods~\cite{sohn2015learning,shen2018improving,gao2019discrete}.

% for their benefits of    Composed of  Add topic model-related literatures. Latent variable models} have been \cmrdy{explored} extensively for dialogue generation~\cite{bowmanetal2016generating}, since it captures the latent semantics while retaining a level of flexibility. This line of work includes hierarchical recurrent VAE methods~\cite{serban2017hierarchical,parketal2018hierarchical,zengetal2019dirichlet} and conditional VAE methods~\cite{sohn2015learning,shen2018improving,gao2019discrete}.

On the other hand, pre-trained language models are used as embedding inputs to VAE-based models~\cite{peineltetal2020tbert,asgari2020topicbert}. Recent work by~\newcite{lietal2020optimus} employs BERT and GPT-2 as the encoder-decoder structure of VAE. However, these models have to be either trained from scratch or built upon pre-trained embeddings. They therefore cannot be directly applied to the low-resource setting of dialogue emotion detection. %, and cannot benefit from the co-occurrence pattern of utterances within dialogues.

% VHRED blablabla On the other side, pre-trained language models are used as complementary representations to enrich with co-occurrences from unlabelled data owing to large amounts of

% Learning approaches can be effective This line of work has became an active field of research
% Our work is also related to latent variable models in dialogue systems.  TopicBert, Optimus. Become the first to use, train from scratch or based on the pre-trained representations,

\paragraph{Knowledge Base and Knowledge Retrieval} %A notable example of a knowledge base is the WordNet~\cite{miller1990introduction}, which made the first attempt to stipulate the word-sense ontology as a knowledge graph. The Google Knowledge Graph~\cite{singhal2012introducing} centers on entities and captures the relations among them. Following a similar vein,
ConceptNet~\cite{speer2017conceptnet} captures commonsense concepts and relations as a semantic network, which encompasses the spatial, physical, social, temporal, and psychological aspects of everyday life. More recently, \newcite{sap2019atomic} built \textsc{Atomic}, a knowledge graph centered on events rather than entities. Owing to the expressiveness of events and ameliorated relation types, using \textsc{Atomic} achieved competitive results against human evaluation in the task of If-Then reasoning.

Alongside the development of knowledge bases, recent years have witnessed the thrive of new methods for training language models from large-scale text corpora as implicit knowledge base. As has been shown in \cite{petroni2019language}, pre-trained language models perform well in recalling relational knowledge involving triplet relations about entities. \newcite{bosselut2019comet} proposed COMmonsEnse Transformers (\textsc{Comet}) which learns to generate commonsense descriptions in natural language by fine-tuning pre-trained language models on existing commonsense knowledge bases such as \textsc{Atomic}. Compared with extractive methods, language models fine-tuned on knowledge bases have a distinctive advantage of being able to generate knowledge for unseen events, which is of great importance for tasks which require the incorporation of commonsense knowledge such as emotion detection in dialogues. % Our work relies on \textsc{Atomic} knowledge base, while taking advantage of the knowledge generator to improve the robustness of knowledge retrieval.

\begin{figure*}[ht]
\centering
% \includegraphics[width=1.0\textwidth]{./figs/fig1}
% \caption{Topic-driven fine-tuning of a pre-trained LM.}
\subfloat[Subfigure 1 list of figures text][Topic-driven fine-tuning of a pre-trained LM.]{
\includegraphics[width=0.48\textwidth]{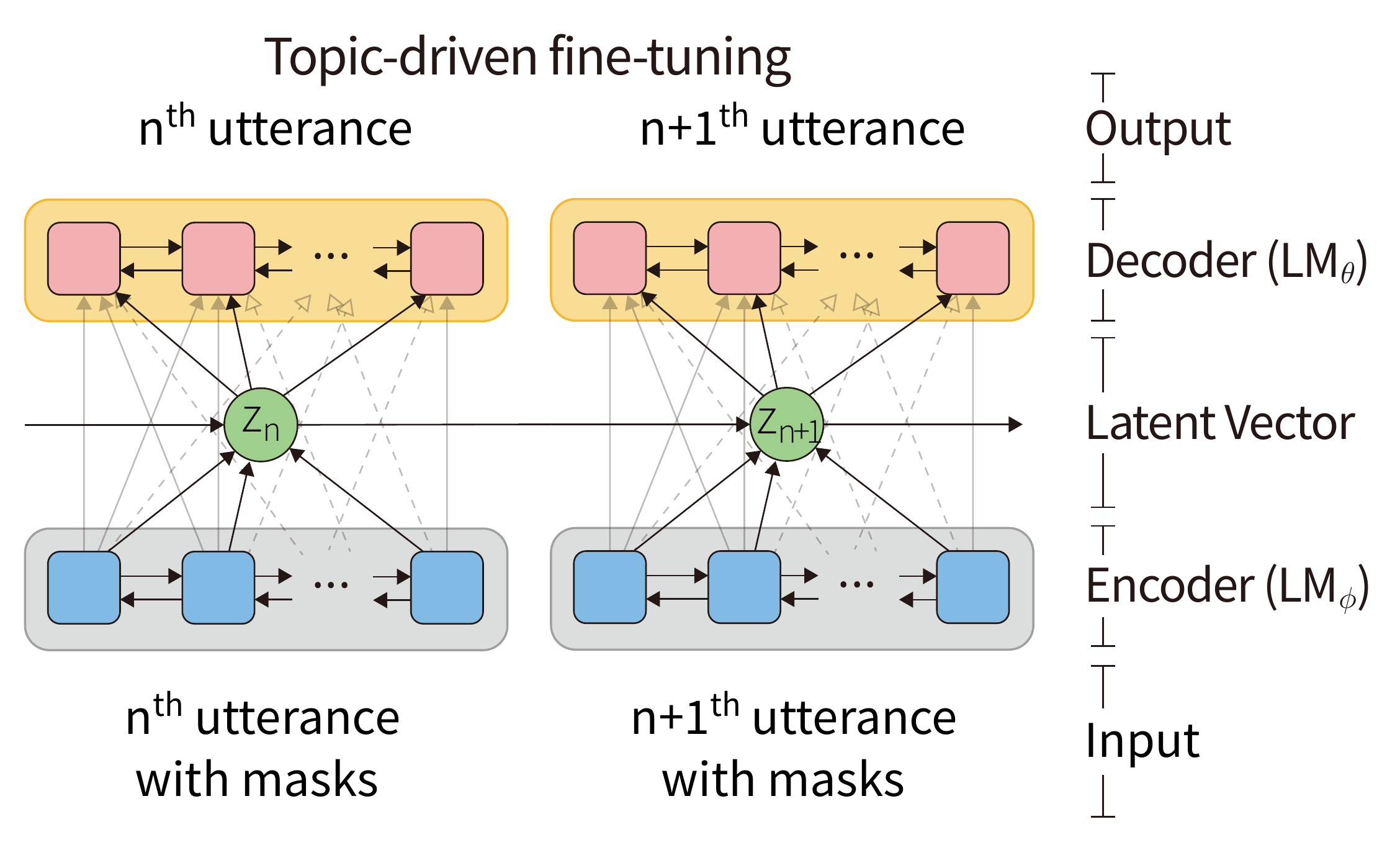}
\label{fig:subfig1d1}}
\subfloat[Subfigure 1 list of figures text][Knowledge-aware transformer.]{
\includegraphics[width=0.48\textwidth]{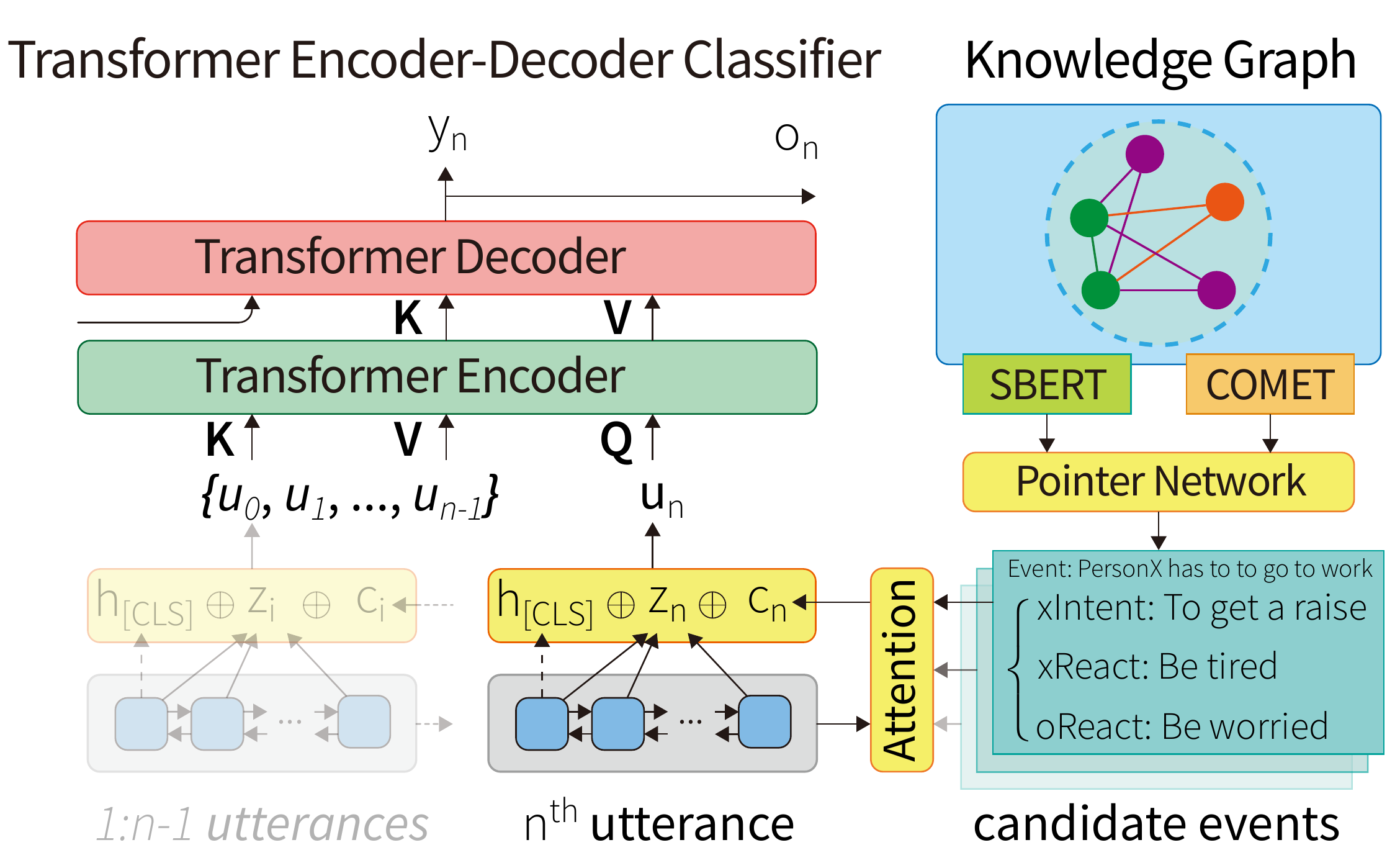}
\label{fig:subfig1d2}}
\caption{TOpic-Driven and Knowledge-Aware Transformer (\textsc{TodKat}).}
\label{fig:1}
\end{figure*}

\section{Methodology}
\subsection{Problem Setup}
A dialogue is defined as a sequence of utterances $\lbrace x_1, x_2, \dots, x_N \rbrace$, which \yh{is annotated with} %corresponds to as well
a sequence of emotion labels $\lbrace y_1, y_2,$ $ \dots, y_N\rbrace$. Our goal is to develop a model that can assign the correct label to each utterance. As for each utterance, the raw input is a token sequence, i.e., $x_n = \lbrace w_{n, 1}, w_{n, 2}, \dots, w_{n, M_n} \rbrace$ where $M_n$ denotes the length of an utterance. We address this problem using the Seq2Seq framework~\cite{sutskever2014sequence}, in which %. In the Seq2Seq framework,
the model consecutively \cmrdy{consumes an utterance $x_n$} and predicts the emotion label $y_n$ based on the earlier utterances %seen so far
and their associated predicted emotion labels. The joint probability of emotion labels for a dialogue is:
% for an utterance $x_n$
\begin{equation}
P_\theta(y_{1:N}|x_{1:N}) = \prod_{n=1}^{N}P_\theta(y_n|x_{\le n}, y_{< n})
\end{equation}
%The task is analogous to machine translation, with the crux that the subsequent utterances are unseen to the model at each predictive step. L
\cmrdy{It is worth mentioning that the subsequent utterances are unseen to the model at each predictive step.} Learning is performed via optimizing the log-likelihoods of \yh{predicted} emotion labels.

The overall architecture of our proposed TOpic-Driven and Knowledge-Aware Transformer (\textsc{TodKat}) is shown in Figure \ref{fig:1}, which consists of two main components, the topic-driven language model fine tuned on dialogues, and the knowledge-aware transformer for emotion label sequence prediction for a given dialogue. In what follows, we will describe each of the components in turn. %In what follows, we first present our topic-driven language model fine-tuning on dialogues. % and discuss the theoretical benefits of our approach.
%Then we present the knowledge-aware transformer \yh{for emotion label sequence prediction for a given dialogue}. %which is the classifier fulfilling the Seq2Seq framework. Consequently, the effectiveness of our approach is demonstrated in Section 4.

% \subsection{Topic-Driven Fine-Tuning of a Language Model}
\subsection{Topic Representation Learning}
We propose to insert a topic layer into an existing language model and fine-tune %The topic-driven fine-tuning is an adaptive tuning method for topic extraction by tuning
the pre-trained language model on the conversational text for topic representation learning. Topic models, %scilicet
often formulated as latent variable models, play a vital role in dialogue modeling~\cite{serban2017hierarchical} due to the explicit modeling of `high-level syntactic features such as style and topic'~\cite{bowman2016generating}. Despite the tremendous success of applying topic modeling in dialogue generation~\cite{sohn2015learning,shen2018improving,gao2019discrete}, there is scarce work exploiting latent variable models for dialogue emotion detection. To this end, we borrow the architecture from VHRED~\cite{serban2017hierarchical} for topic discovery, with the key modification that both the encoder RNN and decoder RNN are replaced by layers of a pre-trained language model. Furthermore, we use a transformer multi-head attention in replacement of the LSTM to model the dependence between the latent topic vectors. Unlike VHRED, we are interested in the encoder part to extract the posterior of the latent topic $z$, rather than the recurrent prior of $z$ in the decoder part since the latter is intended for dialogue generation. \cmrdy{We assume that each utterance is mapped to a latent variable encoding its internal topic, and impose a sequential dependence on the topic transitions.} Figure~\ref{fig:subfig1d1} gives an overview of the VAE-based model which aims %targets
at learning the latent topic vector during the fine-tuning of the language model.
% We no longer employ the shifted-right prediction at the dialogue level since we are not working on dialogue generation, that is, the ground truths are emotion labels instead of the next utterances. However, at the utterance level we still apply the shifted-right scheme to comply with the language models.

Specifically, the pre-trained language model is decomposed into two parts, the encoder and the decoder. By retaining the pre-trained weights, we transfer representations from high-resource tasks to the low-resource setting, which is the case for dialogue emotion datasets.

\paragraph{Encoder} The training of topic discovery part of \textsc{TodKat} comprises a VAE at each time step, with its latent variable dependent on the previous latent code. \cmrdy{Each utterance is input to the VAE encoder with a recurrent hidden state, the output of which is a latent vector ideally encoding the topic discussed in the utterance. The latent vectors are tied through a recurrent hidden state to constraint a coherent topic over a single dialogue.}
%reflect the constraint that they are within the same dialogue.} We use %$x_n^{LM} = \textrm{LM}_{\phi}(x_n)$
We use $\textrm{LM}_{\phi}$ to denote the network of lower layers of the language model (before the topic layer) and $x_n^L$ to denote the output from $\textrm{LM}_{\phi}$ given the input $x_n$. The variational distribution for the approximation of the posterior will be:

{\small
\begin{align}
&\begin{aligned}
&q_\phi(z_n|\bm{x}_{\le n}, \bm{z}_{< n})\\ &=\mathcal{N}\big(z_n|f_{\mu_\phi}(x_n^{L}, h_{n-1}), f_{\sigma_\phi}(x_n^{L}, h_{n-1})\big),\label{eqs:2}
\end{aligned}\\
&\mathrm{where}\  h_{n-1} = f_{\tau}(z_{n-1}, x_{n-1}^{L}) ,\textrm{for} \  n > 1.\label{eqs:3}
\end{align}}
Here, $f_{\mu_\phi}(\cdot)$ and $f_{\sigma_\phi}(\cdot)$ are multi-layer perceptrons (MLPs), $f_\tau$ can be any transition function (e.g., a recurrent unit). We employ the transformer multi-head attention with its query being the previous latent variable $z_{n-1}$, that is,

{\small
\begin{equation}
\label{eqs:4}f_\tau(z_{n-1}, x_{n-1}^{L}) = \mathrm{Attention}(z_{n-1}, x_{n-1}^{L}, x_{n-1}^{L}).
\end{equation}}
\yh{We initialize $h_0 = \mathbf{0}$ and model} %Note that
the transition between $h_{n-1}$ and $h_{n}$ by first generating $z_n$ from $h_{n-1}$ using Eq.~(\ref{eqs:2}), then \yh{calculating $h_n$ by} %performing
Eq.~(\ref{eqs:3}). %$h_0 = \mathbf{0}$.

\paragraph{Decoder} The decoder network reconstructs $x_n$ from $z_n$ at each time step. We use Gaussian distributions for both the generative prior and the variational distribution. Since we want $z_n$ to be dependent on $z_{n-1}$, the prior for $z_n$ is $p(z_n|h_{n-1}) = \mathcal{N}\big(z_n|f_{\mu_{\gamma}}(h_{n-1}),f_{\sigma_{\gamma}}(h_{n-1})\big)$. where $f_{\mu_\gamma}(\cdot)$ and $f_{\sigma_\gamma}(\cdot)$ are MLPs. The posterior for $z_n$ is $p_\theta(z_n|\bm{x}_{\le n}, \bm{z}_{<n})$, which is intractable and is approximated by $q_\phi(z_n|\bm{x}_{\le n}, \bm{z}_{< n})$ of Eq.~\ref{eqs:2}. We denote the higher layers of the language model as $\mathrm{LM}_{\theta}$. Then the \yh{reconstruction} %generation
of $\hat{x}_n$ given $z_n$ and $x_n^{L}$ can be expressed as:
\begin{equation}
    \hat{x}_n = \mathrm{LM}_{\theta}(z_n, x_n^{L}).\label{eq:reconstruction}
\end{equation}
Note that this is different from dialogue generation in which an utterance is generated from the latent topic vector. Here, we aim to extract the latent topic from the current utterance and therefore train the model to reconstruct the input utterance as specified in Eq. (\ref{eq:reconstruction}). %by making $\hat{x}_n$ not entirely relying on $z_n$, our model loses the ability to generate dialogues from the \yh{the latent topic vector}. %prior and decoder.
%However, topics can still be extracted due to the fact that $x_n$ are visible to us before generation, i.e., we are only concerned with the posteriors.
To make the combination of $z_n$ and $x_n^L$ compatible for $\mathrm{LM}_\theta$, we need to perform the latent vector injection. As in~\cite{lietal2020optimus}, we employ the ``\texttt{Memory}'' scheme that $z_n$ becomes an additional input for $\mathrm{LM}_\theta$, that is, the input to the higher layers becomes $[z_n, x_n^{L}]$.

\paragraph{Training} The training objective is the Evidence Lower Bound (ELBO):
\begin{equation}
\begin{aligned}
\label{eqs:6}\mathbb{E}_{q_\phi(\bm{z}_{\le N}|\bm{x}_{\le N})}[\mathrm{log}\;p_\theta(\bm{x}_{\le N}|\bm{z}_{\le N})]\\
- \mathrm{KL}[q_\phi(\bm{z}_{\le N}|\bm{x}_{\le N})||p(\bm{z}_{\le N})].
\end{aligned}
\end{equation}
\cmrdy{Eq.~\ref{eqs:6} factorizes and the expectation term becomes
\begin{equation}
\mathbb{E}_{q_\phi(\bm{z}_{\le N}|\bm{x}_{\le N})}\left[ \sum_{n=1}^{N}\mathrm{log}\;p_\theta(x_{n}|\bm{z}_{\le n}, \bm{x}_{< n})\right],
\end{equation}
and the KL term becomes
\begin{equation}
\sum_{n=1}^{N}\mathrm{KL}[q_\phi(z_{n}|\bm{x}_{\le n}, \bm{z}_{< n})||p(z_n|\bm{z}_{< n},\bm{x}_{< n})],
\end{equation}
where $p(z_n|\bm{z}_{< n},\bm{x}_{< n})$ is the prior for $z_n$. }
After training, we are able to extract the topic representation from the encoder part of the model, which is denoted as $z_n=\mathrm{LM}_\phi^{\mathrm{enc}}(x_n)$. Meanwhile, the entire language model has been fine-tuned, which is denoted as $u_n=\mathrm{LM}^{\mathrm{CLS}}(x_n)$.

\subsection{Knowledge-Aware Transformer}
The topic-driven LM fine-tuning \yh{stage makes it possible for the LM} %equips the language models with the capability
to discover a topic representation from a given utterance. After fine-tuning, we attach the fine-tuned components to a classifier and train the classifier to predict the emotion labels. We propose to use the Transformer Encoder-Decoder structure as the classifier, \yh{and consider the incorporation of commonsense} %. We also consider augmenting the utterances with
knowledge retrieved from \yh{external} knowledge \yh{sources}. %graph, via a pointer network.
In what follows, we first describe how to retrieve the \yh{commonsense} knowledge from a knowledge source, then we present the detailed structure of the classifier.

\paragraph{Commonsense Knowledge Retrieval} We use \textsc{Atomic}\footnote{\url{https://homes.cs.washington.edu/~msap/atomic/}} as a source of external knowledge. In \textsc{Atomic}, each node is a phrase describing an event. Edges are relation types linking from one event to another. \textsc{Atomic} thus encodes triples such as $\langle$\texttt{event, relation type, event}$\rangle$. There are a total of nine relation types, of which three are used: \texttt{xIntent}, the intention of the subject (e.g., `\emph{to get a raise}'), \texttt{xReact}, the reaction of the subject (e.g., `\emph{be tired}'), and \texttt{oReact}, the reaction of the object (e.g., `\emph{be worried}'), since they are defined as the mental states of an event~\cite{sap2019atomic}.

Given an utterance $x_n$, we can compare it with every node in the knowledge graph, and retrieve the most similar one. %Here, the top-$K$ most similar events are extracted from \textsc{Atomic}.
The method for computing the similarity between an utterance and events is SBERT~\cite{reimers2019sentence}. We extract the top-$K$ events, and obtain their intentions and reactions, which are denoted as $\lbrace e^{sI}_{n,k}, e^{sR}_{n,k}, e^{oR}_{n,k} \rbrace, k=1, \dots, K$.

On the other hand, \yh{there is a knowledge generation model, called} \textsc{Comet}\footnote{\url{https://github.com/atcbosselut/comet-commonsense}}, \yh{which is trained on \textsc{Atomic}. It} can \yh{take} %consume
$x_n$ \yh{as input} and \yh{generate} %spawn
the knowledge %(e.g., the \emph{intention} or the \emph{reaction})
with the desired \yh{event relation types} specified (e.g., \texttt{xIntent}, \texttt{xReact} or \texttt{oReact}). The \yh{generated} knowledge can \yh{be unseen in \textsc{Atomic}} %even be non-existent
since \textsc{Comet} is essentially a fine-tuned language model. We use \textsc{Comet} to generate the $K$ most likely events, each with respect to the three \yh{event relation types}. %dimensions.
The produced events are denoted as $\lbrace g^{sI}_{n,k}, g^{sR}_{n,k}, g^{oR}_{n,k} \rbrace$, $ k=1, \dots, K$.

\paragraph{Knowledge Selection} With the knowledge retrieved from \textsc{Atomic}, we build a pointer network~\cite{vinyals2015pointer}
to exclusively choose the commonsense knowledge either from SBERT or \textsc{Comet}. The pointer network calculates the probability of choosing the candidate knowledge source as:
\begin{equation}
\begin{aligned}
P\big(\mathbb{I}(x_n, \bm{e}_n,\bm{g}_n)=1\big)
= \sigma\big(\lbrack x_n, \bm{e}_n, \bm{g}_n \rbrack \mathbf{W}_\sigma\big), \nonumber
\end{aligned}
\end{equation}
where $\mathbb{I}(x_n, \bm{e}_n,\bm{g}_n)$ is an indicator function with value $1$ or $0$, and $\sigma(x)=1/(1+\mathrm{exp}(-x))$. We envelope $\sigma$ with Gumbel Softmax~\cite{jang2016categorical} to generate the one-hot distribution\footnote{We have also experimented with a soft gating mechanism by aggregating knowledge from SBERT and \textsc{Comet} in a weighted manner. But the results are consistently worse than those using a hard gating mechanism.}. The integrated commonsense knowledge is expressed as
\begin{equation}
\bm{c}_n = \mathbb{I}(x_n, \bm{e}_n, \bm{g}_n)\bm{e}_n + \big(1 - \mathbb{I}(x_n, \bm{e}_n, \bm{g}_n)\big)\bm{g}_n, \nonumber
\end{equation}
where $\bm{c}_n = \lbrace c_{n,k}^{sI}, c_{n,k}^{sR}, c_{n,k}^{oR} \rbrace_{k=1}^{K}$.

With the knowledge source selected, we proceed to select the most informative knowledge. We design an attention mechanism~\cite{bahdanau2015neural} to integrate the candidate knowledge. Recall that we have a fine-tuned language model which can calculate both the \texttt{[CLS]} and topic representations. Here we apply the language model to the \yh{retrieved or generated} knowledge  to obtain the \texttt{[CLS]} and the topic representation, denoted as $[\bm{c}_{n,k},z_{n,k}]$. The attention mechanism is performed by calculating the dot product between the utterance and each normalized knowledge tuple:
\begin{gather}
    v_{k} = \mathrm{tanh}\big([\bm{c}_{n,k}, z_{n,k}]\mathbf{W}_\alpha\big), \label{eqs:7}\\ %\nonumber\\
    \alpha_{k} = \frac{\mathrm{exp}\big(v_{k}[z_n, u_n]^\top\big)}{\sum_{k} \mathrm{exp}\big(v_{k}[z_n, u_n]^\top\big)}, \\
    \bm{c}_{n} = \sum_{k=1}^{K}\alpha_{k}\bm{c}_{n,k}.
\end{gather}
Here, we abuse $\bm{c}_n$ to represent the aggregated knowledge phrases. We further aggregate $\bm{c}_n$ by event relation types using a self-attention and the final event representation is denoted as $c_n$.
% Finally, a self-attention is applied to $\bm{c}_n$ to select the most informative event type. The final knowledge representation is denoted as $c_n$.

% And the most informative knowledge is denoted as As for the event types we apply the self-attention to obtain

% If you choose to use

\paragraph{Transformer Encoder-Decoder} We use a Transformer encoder-decoder to map an utterance sequence to an emotion label sequence, thus allowing for modeling the transitional patterns between emotions and taking into account the historical utterances as well. Each utterance is converted to the \texttt{[CLS]} representation concatenated with the topic representation $z_n$ and knowledge representation $c_n$. \cmrdy{We enforce a masking scheme in the self-attention layer of the encoder to make the classifier predict emotions in an auto-regressive way, entailing that only the past utterances are visible to the encoder. This masking strategy, preventing the query from attending to future keys, suits better a real-world scenario in which
%is more natural due to the fact that
the subsequent utterances are unseen when predicting an emotion of the current utterance.
}
% There is a masking strategy in the self-attention sub-layer of encoder to prevent utterances from attending to subsequent ones, since the latter are unseen to the classifier when making the decision on the $n$-th utterance.
As for the decoder, the output of the previous decoder block is input as a query to the self-attention layer. The training loss for the classifier is the negative log-likelihood expressed as:
\begin{equation}
    \mathcal{L} = - \sum_{n=1}^{N} \mathrm{log}\ p_\theta(y_n|\bm{u}_{\le n}, \bm{y}_{< n}), \nonumber
\end{equation}
where $\theta$ denotes the trainable parameters.

\section{Experimental Setup}
In this section, we present the details of the datasets used, the methods for comparison, and the implementation details of our models.

\paragraph{Datasets} We use the following datasets for experimental evaluation:

\noindent\underline{DailyDialog}~\cite{li2017dailydialog} is collected from daily communications. It takes the  Ekman's six emotion types~\cite{ekman1993facial} as the annotation protocol, that is, it annotates an utterance with one of the six basic emotions: \emph{anger, disgust, fear, happiness, sadness}, or \emph{surprise}. Those showing ambiguous emotions are annotated as \emph{neutral}.

\noindent\underline{MELD}~\cite{poria2019meld} is constructed from scripts of `\emph{Friends}', a TV series on urban life. Same as DailyDialog, the emotion label falls into Ekman’s six emotion types, or \emph{neutral}.

\noindent\underline{IEMOCAP}~\cite{busso2008iemocap} is built with subtitles from improvised videos. Its emotion labels are \emph{happy, sad, neutral, angry, excited} and \emph{frustrated}.

\noindent\underline{EmoryNLP}~\cite{zahiri2018emotion}\footnote{\url{https://github.com/emorynlp/emotion-detection}} \cmrdy{is also built with conversations from `\emph{Friends}' TV series, but with a slightly different annotation scheme in which \emph{disgust, anger} and \emph{surprise} become \emph{peaceful, mad} and \emph{powerful}, respectively.}
%can be viewed as a version of trimmed MELD. % It uses the same annotation scheme\cmrdy{, with slightly different from the }.

\cmrdy{Following \citet{zhong2019knowledge} and \citet{ghosal2020cosmic}, the `\emph{neutral}' label of DailyDialog is not counted in the evaluation to avoid highly imbalanced classes.} For MELD and EmoryNLP, we consider a dialogue as a sequence of utterances from the same scene ID. Table~\ref{tab:1} summarizes the statistics of each dataset.

% The statistics of the datasets are summarized in Table

% related knowledge
% eccentric

\begin{table}[htp]
\begin{center}
\fontsize{9pt}{11pt}\selectfont
\begin{tabular}{l|cccc}
\thickhline
\hline
&  DD &  MELD &  IEMOCAP &  EmoryNLP \\ \hline
\#Dial. & 13,118 & 1,432 & 151 & 827 \\
\, Train & 11,118 & 1,038 & 100 & 659 \\
\, Dev. & 1,000 & 114 & 20 & 89 \\
\, Test & 1,000 & 280 & 31 & 79 \\ \hline
\#Utt.& 102,979 & 13,708 & 7,333 & 9,489\\
\, Train & 87,170 & 9,989 & 4,810 & 7,551 \\
\, Dev. & 8,069 & 1,109 & 1,000 & 954 \\
\, Test & 7,740 & 2,610 & 1,523 & 984 \\ \hline
% Utterances & 87,170/8,069/7,740 & 9,989/1,109/2,610 & 4,810/1,000/1,523 & 7,551/954/984 \\
\#Cat. & 7 & 7 & 6 & 7\\
\thickhline
\end{tabular}
\end{center}
\caption{\label{tab:1} Statistics of the benchmarks for dialogue emotion detection. The train/development/test sets are pre-defined in each dataset.} %Every benchmark has provided a training set, a development set and a testing set, which is detailed in the number of utterances.}
\end{table}

\renewcommand{\thefootnote}{\fnsymbol{footnote}}

\begin{table*}[htp]
\begin{center}
\fontsize{9pt}{11pt}\selectfont
\begin{tabular}{l|l|l|l|l|l|l|l|l}
\thickhline
\hline
\multicolumn{1}{l|}{\multirow{3}{*}{Models}} &  \multicolumn{2}{c|}{DailyDialog} &  \multicolumn{2}{c|}{MELD} &  \multicolumn{2}{c|}{IEMOCAP} &  \multicolumn{2}{c}{EmoryNLP} \\
\cline{2-9}
& \multicolumn{1}{c|}{\multirow{2}{*}{\tabincell{c}{Macro-F1\\ - neutral}}} & \multicolumn{1}{c|}{\multirow{2}{*}{\tabincell{c}{Micro-F1\\ - neutral}}} & \multicolumn{1}{c|}{\multirow{2}{*}{\tabincell{c}{weighted \\ Avg-F1}}} & \multicolumn{1}{c|}{\multirow{2}{*}{Micro-F1}} & \multicolumn{1}{c|}{\multirow{2}{*}{\tabincell{c}{weighted\\ Avg-F1}}} & \multicolumn{1}{c|}{\multirow{2}{*}{Micro-F1}} & \multicolumn{1}{c|}{\multirow{2}{*}{\tabincell{c}{weighted\\ Avg-F1}}} & \multicolumn{1}{c}{\multirow{2}{*}{Micro-F1}} \\
& & & & & & & &\\
\hline
HiGRU & 0.4904 & 0.5190 & 0.5681 & 0.5452 & 0.5854 & 0.5828 & 0.3448 & 0.3354\\
DialogueGCN & 0.4995 & 0.5373 & 0.5837 & 0.5617 & 0.6085 & 0.6063 & 0.3429 & 0.3313\\
KET & -- & 0.5348 & 0.5818 & --  & 0.5956 & -- & 0.3439 & --\\
COSMIC & 0.5105 & \textbf{0.5848} & 0.6521 & -- & \textbf{0.6528}\footnote[1]{This is incomparable due to the larger training set.} & -- & 0.3811 & --\\
\hline
\hline
\textsc{TodKat} & \textbf{0.5256} & 0.5847 & \textbf{0.6823} & \textbf{0.6475} & 0.6133 & 0.6111 & \textbf{0.4312} & \textbf{0.4268}\\
\quad $-$Topics & 0.5136 & 0.5549 & 0.6634 & 0.6352 & 0.6281 & \textbf{0.6260} & 0.4180 & 0.4055\\
\quad $-$KB& 0.5003 & 0.5344 & 0.6397 & 0.6111 & 0.5896 & 0.5738 & 0.3379 & 0.3262\\
% \quad \textsc{Kat} + ConceptNet & 0.5407 & 0.5825 & 0.5970 & 0.3957\\
\quad \textsc{Kat}\textsubscript{SBERT} & 0.5173 & 0.5578 & 0.6454 & 0.6188 & 0.6097 & 0.6069 & 0.3734 & 0.3567\\
\quad \textsc{Kat}\textsubscript{\textsc{Comet}} & 0.5102 & 0.5462 & 0.6582 & 0.6307 & 0.6277 & 0.6254 & 0.4110 & 0.3974\\
\thickhline
\end{tabular}
\end{center}
\caption{\label{tab:2} The F1 results of the dialogue emotion detectors on four benchmarks. Here we denote the proposed model as \textsc{TodKat}, of which the results are an average of ten runs. The ablations of different components are reported separately in the bottom, where the \yh{model without the incorporation of latent topics is denoted as} `$-$Topics', transformer encoder-decoder structure without the use of a knowledge base is dnoted as `$-$KB'. \textsc{Kat}\textsubscript{\textsc{Comet}} and \textsc{Kat}\textsubscript{SBERT} uses the commonsense knowledge obtained with \textsc{Comet} and SBERT, respectively. Results of KET and COSMIC are from \cite{zhong2019knowledge} and \cite{ghosal2020cosmic}, respectively.}
% knowledge only. `\textsc{Kat} with \textsc{Atomic} Extractor' and `\textsc{Kat} with \textsc{Atomic} Generator' use the commonsense knowledge obtained from \textsc{Atomic} using the Knowledge Extractor and the Knowledge Generator, respectively.} Results of baseline methods are taken from
\end{table*}
% with clear leadership
% Visualize the attention score / weights towards
%\renewcommand*{\thefootnote}{\arabic{footnote}}
\renewcommand{\thefootnote}{\arabic{footnote}}

\paragraph{Baselines} We compare the performance of \textsc{TodKat} with the following methods:

\noindent\underline{HiGRU}~\cite{jiao2019higru} simply inherits the recurrent attention framework that an attention layer is placed between two GRUs to aggregate the signals from the encoder GRU and pass them to the decoder GRU.

% The model has been evaluated on the IEMOCAP dataset, however with only three emotions (\emph{anger, happiness, sadness}) kept and other emotions abandoned.
% omitted.

\noindent\underline{DialogueGCN}~\cite{ghosal2019dialoguegcn} creates a graph from interactions of speakers to take into account the dialogue structure. A Graph Convolutional Network (GCN) is employed to encode the speakers. Emotion labels are predicted with the combinations of the global context and speakers' status.

% Since we do not distinguish between the speakers' identifications, we regard all dialogues as produced by two speakers to meet the requirements of the model.
% we manually  emphasize the role of speakers. It employs an GCN to  identifies the speakers and encodes their interactions using an GCN.

\noindent\underline{KET}~\cite{zhong2019knowledge} is the first model which integrates common-sense knowledge extracted from ConceptNet and emotion information from an emotion lexicon into conversational text. \cmrdy{A Transformer encoder is employed to handle the influence from past utterances.}

\noindent\underline{COSMIC}~\cite{ghosal2020cosmic} is the state-of-the-art approach that leverages \textsc{Atomic} for improved emotion detection. \textsc{Comet} is employed in their model to retrieve the event-eccentric commonsense knowledge %phrases
from \textsc{Atomic}. %the knowledge base.

We modified the script\footnote{\url{https://huggingface.co/transformers/v2.0.0/examples.html}} of language model fine-tuning \yh{in the} Hugging Face library~\cite{wolfetal-2020-transformers} for the implementation of topic-driven fine-tuning. \cmrdy{We use one transformer encoder layer. As for the decoder, there are $N$ layers where $N$ is the number of utterances in a dialogue.}
We refer the readers to the Appendix for the detailed settings of the proposed models.

\section{Results and Analysis}
\paragraph{Comparison with Baselines} Experiment results of \textsc{TodKat} and its ablations are reported in Table~\ref{tab:2}. HiGRU and DialogueGCN results were produced by running the code published by the authors on the four datasets. % while KET results were taken from %For KET, we present the results of
% \cite{zhong2019knowledge}.
\yh{Among the baselines, COSMIC gives the best results. Our proposed \textsc{TodKat} outperforms COSMIC on both MELD and EmoryNLP in weighted Avg-F1 with the improvements ranging between 3-5\%. \textsc{TodKat} also achieves superior result than COSMIC on DailyDialogue in Macro-F1 and gives nearly the same result in Micro-F1.} %We observe consistent improvements over the baselines across datasets. \textsc{TodKat} exceeds COSMIC on six metrics, while in micro-F1 metric it was $0.01\%$ slightly surpassed by COSMIC on DailyDialog.
\textsc{TodKat} is inferior to COSMIC on IEMOCAP. It is however worth mentioning that COSMIC was trained with $132$ instances on this dataset, while for all the other models the training-and-validation split is $100$ and $20$. \yh{As such, the IEMOCAP results reported on COSMIC \cite{ghosal2020cosmic} are not directly comparable here. COSMIC also incorporates the commonsense knowledge from \textsc{Atomic} but with the modified GRUs. Our proposed \textsc{TodKat}, built upon the topic-driven Transformer, appears to be a more effective architecure for dialogue emotion detection.} Compared with KET, the improvements are much more significant, with over $10\%$ increase on MELD, and close to $5\%$ gain on DailyDialog. KET is also built on the Transformer, but it considers each utterance in isolation and applies commonsense knowledge from ConceptNet. \textsc{TodKat}, on the contrary, takes into account the dependency of previous utterances and their associated emotion labels for the prediction of the emotion label of the current utterance. %This might be credited to the RoBERTa embedding, while it can't be denied that commonsense knowledge also plays an important part since the `\textsc{Kat} - KB' is not as good as KET.
DialogueGCN models interactions of speakers and it performs slightly better than KET. But it is significantly worse than \textsc{TodKat}. It seems that %, which supports our claim that
topics might be more useful in capturing the dialogue context.

\paragraph{Ablation Study} \yh{The lower half of } Table~\ref{tab:2} presents the F1 scores with the removal of various components from \textsc{TodKat}. It can be observed that with the removal of the topic component, the performance of \textsc{TodKat} drops consistently across all datasets except IEMOCAP in which we observe a slight increase in both weighted average F1 and Micro-F1.
%\textsc{TodKat} achieves the best overall performance, however with weighted average-F1 and micro-F1 slightly exceeded by removing the topic component (`$-$Topics') and \textsc{Kat}\textsubscript{\textsc{Comet}} on IEMOCAP.
This might be attributed to the size of the data since IEMOCAP is the smallest dataset evaluated here\cmrdy{, and small datasets hinder the model's capability to discover topics.} % Also using commonsense knowledge generated from \textsc{Comet} appears to be more effective on IEMOCAP.} %This leads to the conclusion that topic-driven fine-tuning is effective in most circumstances. While some exception could happen as in IEMOCAP possibly due to the rich commonsense knowledge improvised by the players. To examine the effects of different methods for commonsense knowledge extraction,
Without using the commonsense knowledge (`$-$KB'), we observe more drastic performance drop compared to all other components, with nearly 10\% drop in F1 on EmoryNLP, showing the importance of employing commonsense knowledge for dialogue emotion detection. Comparing two different ways of extracting knowledge from \textsc{Atomic}, direct retrieval using \textsc{Sbert} or generation using \textsc{Comet}, we observe mixed results. %we remove the knowledge extractors step by step. There is an advantage of the \textsc{Comet} on datasets like MELD, IEMOCAP and EmoryNLP, while on DailyDialog it is outperformed by SBERT.
Overall, the Transformer Encoder-Decoder with a pointer network is a conciliator between the two methods, yielding a balanced performance across the datasets. %\textsc{Kat} performs the worst in all the datasets.

\paragraph{Relationships between Topics and Emotions}
To investigate the effectiveness of the learned topic vectors, we perform t-SNE~\cite{van2008visualizing} on the test set to study the relationship %correspondence
between the learned topic vectors and the ground-truth emotion labels. %We
\yh{The results on DailyDialog and MELD are} illustrated %the correlation
in Figure~\ref{fig:3}(a) and (b). %Hidden variables
\yh{Latent topic vectors} of utterance are used to plot the data points, whose colors indicate their \yh{ground-truth} emotion labels. We can see that the majority of the topic vectors cluster into polarized groups. Few clusters are bearing a mixture of polarity, possibly due to the background topics such as greetings in the datasets. %The clustering property indicates that topics are polarized, which are helpful for classifying the emotion labels.

\definecolor{forestgreen}{RGB}{27,111,27}

\renewcommand{\floatpagefraction}{.7}%

\begin{figure}[h!]
\begin{minipage}{.238\textwidth}
\centering
\subfloat[Subfigure 1 list of figures text][DailyDialog]{
\includegraphics[scale=.222]{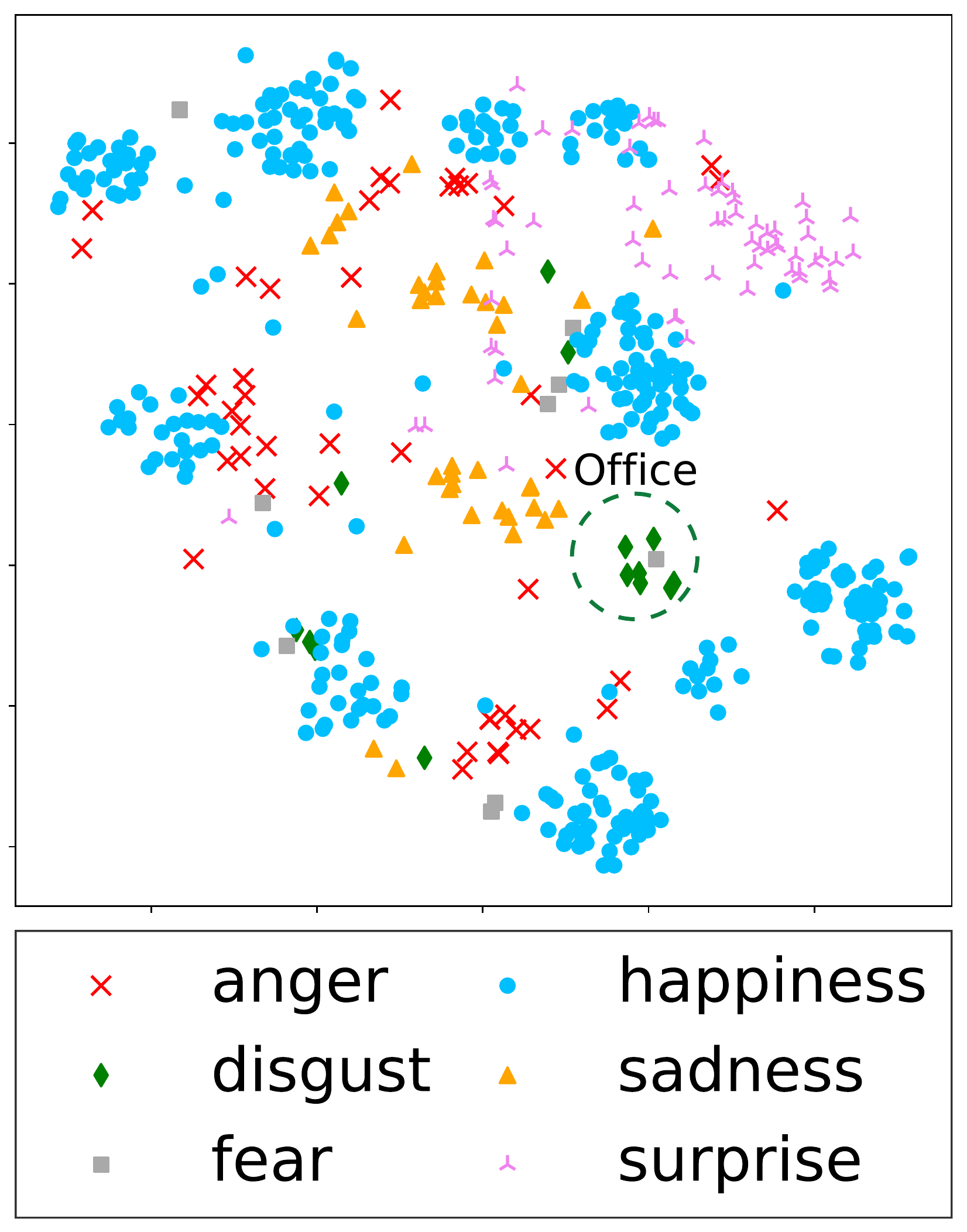}
\label{fig:subfig1}}
\end{minipage}
\begin{minipage}{.238\textwidth}
\centering
\subfloat[Subfigure 2 list of figures text][MELD]{
\includegraphics[scale=.223]{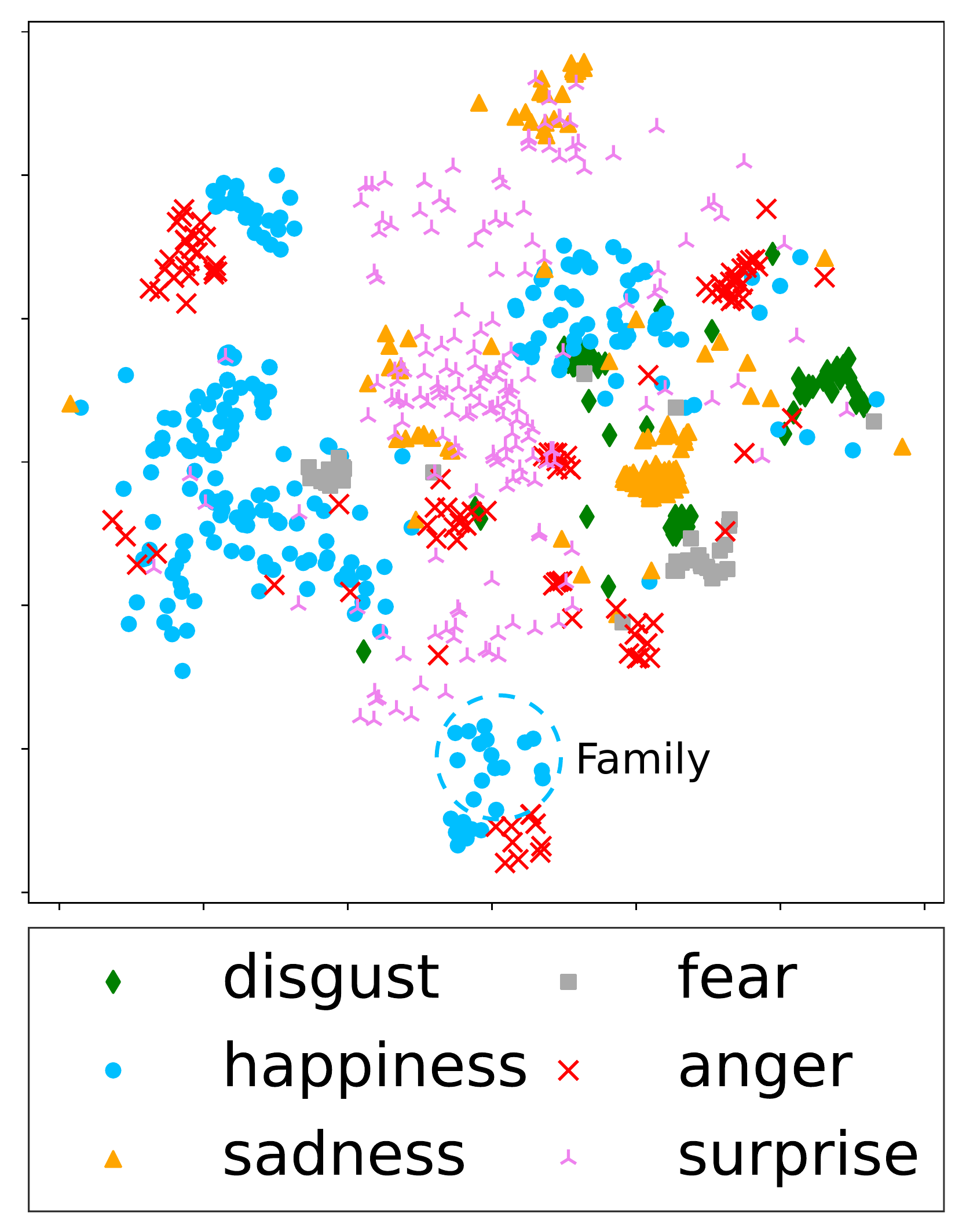}
\label{fig:subfig2}}
\end{minipage}
\centering
\subfloat[Subfigure 3 list of figures text][Representative utterances and their topics]{
\fontsize{8pt}{10pt}\selectfont
\begin{tabular}{l|lc}
\thickhline
\hline
Topic & \multicolumn{1}{c}{Utterances} &  \makebox[\dimexpr(\width-0.5em)][r]{Emotion}\\
\hline
% \tabincell{l}{Office\\Business\\Vacation}
\tabincell{l}{Office}& \tabincell{l}{A: How are you doing, Christopher?\\ \textcolor{forestgreen}{B: To be honest, I'm really fed up with}\\ \quad\, \textcolor{forestgreen}{work at the moment. I need a break!}\\ A: Are you doing anything this weekend?\\ \textcolor{forestgreen}{B: I have to work on Saturday all day!}\\ \quad\, \textcolor{forestgreen}{I really hate my job!}} & \tabincell{c}{\makebox[\dimexpr(\width-0.5em)][r]{\textcolor{forestgreen}{disgust}}}\\
%\makebox[\dimexpr(\width-0.5em)][r]{\textcolor{forestgreen}{disgust}}\\ \\ neutral \\ disgust \\ \\}\\
\hline
Family & \tabincell{l}{\textcolor{cyan}{A: Yeah, I-I heard. I think it's great! Ohh, }\\
\quad\, \textcolor{cyan}{I'm so happy for you!}\\
\textcolor{cyan}{B: I can't believe you're getting married!}\\
\textcolor{black}{C: Yeah.}\\
\textcolor{black}{D: Monica and Rachel made out.}} & \makebox[\dimexpr(\width-0.5em)][r]{\textcolor{cyan}{happy}}\\
%\quad\, \textcolor{cyan}{--friends of ours were in the audience,}\\ \quad\, \textcolor{cyan}{..., like family members.}\\ \textcolor{cyan}{B: Yeah. It was great.}\\ \textcolor{cyan}{A: It was--I cried,}}

%Family & \tabincell{l}{A: He asked me to marry him, like- onstage. \\ \quad\, I was like, I couldn't believe.\\ A: Yeah, I said yes, but I, yeah, I mean,\\ A: I don't know- it just like took me, \\ \quad\, took me by surprise.\\ B: He scored points.\\ A: Oh, yeah. Everyone--yeah. And people\\ \quad\, --friends of ours were in the audience,\\ \quad\, I didn't know about, like family members.\\ B: Yeah. It was great.\\ A: It was--I cried,} & \textcolor{cyan}{happy}\\
\thickhline
\end{tabular}
\label{fig:subfig3}}
\caption{%T-SNE visualization on DailyDialog and MELD.
% Different colors indicate different emotions.
T-SNE visualization of the learned topic vectors of utterances from the test sets of DailyDialog (subfigure (a)) and MELD (subfigure (b)). Colors indicate the ground-truth emotion label. Neutral utterances are omitted here for clarity. Representative utterances (highlighted in colors) for the topic `\emph{Office}' in DailyDialog and the topic `\emph{Family}' in MELD are shown in subfigure (c). } %with the same colour have the same emotion label as shown in the last column. Visualization and highlight of the neutral utterances are omitted for clarity}. Each cluster is exemplified by a group of utterances.}
% \caption{Topic-Emotion t-sne on DailyDialog. grey: fear, green: disgust, blue: happiness, orange: sadness, red: anger, purple: surprise. Topic-Emotion t-sne on IEMOCAP. grey: neutral, green: frustrated, blue: happy, orange: sad, red: angry, purple: excited.}
% Interrupted corrupted vacation
\label{fig:3}
\end{figure}

% \renewcommand{\floatpagefraction}{.5}%
% conjecture

Topics can be interpreted using the attention scores of Eq.~\ref{eqs:4}. The top-10 most-attended words are selected as the representative words for each utterance. As in~\cite{Dathathri2020Plug}, we construct bag-of-words\footnote{Word lists and their corresponding theme names are crawled from \url{https://www.enchantedlearning.com/wordlist/}.} that represent $141$ distinct topics. Given the attended words of an utterance \yh{cluster grouped based on their latent topic representations}, %collection,
we %use the word list to filter out stop words and
label the word collection with the dominant theme name. We refer to the theme names as topics in Figure~\ref{fig:subfig3}. \cmrdy{It can be observed that utterances associated with Office tend to carry \emph{`disgust'} emotions, while those related to \emph{Family} are prone to be \emph{`happy'}. }
%There are also cases where similar utterances exhibit different emotions due to their associated topics, e.g., }
%A: Johnny died yesterday, we knew that it was coming, but... \\
%B: Like just last week, he was doing so well.
%and
%A: Then all of a sudden they gave him a microphone, he asked me to marry him, like, onstage. \\
%B: He scored points.

%, showing that the emotion of interlocutors heavily depends on the topics they are talking about. % Usually topics play a major part in determining the emotion, but emotion transition also contributes to the emotion changes.}
% non-informative
% thanks to the
% Following
% attention weights, scores
% validates
% indicates that
% As in~\cite{lietal2020optimus}
% Additionally
% correlation
% We further perform. We show a heat map between topics and emotion vectors in Figure . validate
% observed a significant

%We notice an overlap among the emotion groups in Figure~\ref{fig:subfig2}. In this concern,
We further compute the Spearman's rank-order correlation coefficient to quantitatively verify the relationship between the topic and emotion vectors.
% cosine-similarities of topic vectors and that of
For an utterance pair, a similarity score is obtained separately for their corresponding topic vectors as well as their emotion vectors. %from each topic vector pair by calculating their cosine similarity. A similar score is also obtained from the corresponding emotion vector pair.
We \cmrdy{then sort the list of emotion vector pairs according to their similarity scores to check to what extent their ranking matches
%see how its order coordinates with
that of topic vector pairs, based on the Spearman’s rank-order correlation coefficient.}
%compute the Spearman's rank-order correlation coefficient between these two lists of scores for all utterance pairs in the test set.
The results are $0.60, 0.58, 0.42$ and $0.54$ with p-values $\ll 0.01$  respectively for DailyDialog, MELD, IEMOCAP and EmoryNLP, showing that there is a strong correlation between the clustering of topics and that of emotion labels. IEMOCAP has the lowest correlation score, which is inline with the results in Table 2 that the discovered latent topics did not improve the emotion classification results. %This might also explain why incorporating topics hinders the performance on IEMOCAP  since the discovered topics on IEMOCAP entangled and mixed up the emotions.

% Aka, the input.
% has convinced us of the necessity of topic modelling.
% Conclusively, statistically conclude that, confirm

% (How to compute them, sample). The spear, showing that the is prominent , this has convinced us that is indeed for

% Topics are a good choice to select knowledge. In the feature, we will explore fine-tuning knowledge language model on the downstream task.
% Related works. The OPTIMUS, VHRED, diaolgue emotion detection especially those knowledge-based,
% The
% between cosine-similarities of topic vectors and that of emotion vectors.

% To investigate the effectiveness of topic vectors, we compute the Spearman rank-order correlation coefficient between cosine-similarities of topic vectors and that of emotion vectors.

% \paragraph{Visualization of the Latent Space}
% As in~\cite{lietal2020optimus},

% To demonstrate the effectiveness of commonsense knowledge exploited, we examine the take away of knowledge bases. The results show that

% accredit for

% There might be some errors that attribute to the
\begin{table}[htb]
\begin{center}
\fontsize{9pt}{11pt}\selectfont
\begin{tabular}{c|cc}
\thickhline
\hline
\multicolumn{1}{c|}{\multirow{2}{*}{Dataset}} &  \multicolumn{2}{c}{Relation Type} \\ \cline{2-3}
 & $\{sI, sR, oR, sE, oE\}$ & All\\
\hline
DailyDialog & 0.5718$\downarrow$ & 0.5664$\downarrow$ \\
MELD & 0.6429$\downarrow$ & 0.6322$\downarrow$ \\
IEMOCAP & 0.6163$\uparrow$ & 0.6073$\downarrow$ \\
EmoryNLP & 0.4029$\downarrow$ & 0.3885$\downarrow$ \\
\thickhline
\end{tabular}
\end{center}
\caption{\label{tab:3} Micro-F1 scores of \textsc{TodKat} with more commonsense relation types retrieved from \textsc{Atomic} included for training. Here, ``$sE$'' and ``$oE$'' represent \emph{effect of subject} and \emph{effect of object}, respectively. ``All'' denotes the incorporation of all nine commonsense relation types from \textsc{Atomic}.}
\end{table}
\paragraph{Impact of Relation Type}
We investigate the impact of commonsense relation types on the performance of \textsc{TodKat}. We expand the relation set to five relation types and all nine relation types, respectively. According to~\cite{sap2019atomic}, there are other relation types including $\{\emph{sNeed}, \emph{sWant}, \emph{oWant}, \emph{sEffect}, \emph{oEffect}\}$, which identifies the prerequisites and post conditions of the given event, and $\{\emph{sAttr}\}$, the ``If-Event-Then-Persona'' category of relation type that describes how the subject is perceived by others. We calculate the Micro-F1 scores of \textsc{TodKat} with these two categories of relation types added step by step. From Table~\ref{tab:3} we can conclude that the inclusion of two extra relation types or all relation types %some relations, which are categorized as ``If-Event-Then-Event'' and ``If-Event-Then-Persona'' by \newcite{sap2019atomic},
degrades the F1 scores on almost all datasets. An exception occurs on IEMOCAP where the F1 score rises by $0.5\%$ when adding ``$sE$'' and ``$oE$'' relations, possibly due to the fact that the dataset is abundant in events. Hence the extra event descriptions offer complementary knowledge to some extent. While on other datasets neither the incorporation of ``If-Event-Then-Event'' nor the incorporation of ``If-Event-Then-Persona'' relation types could bring any benefit.

% the effectiveness in
% in the knowledge selection

\paragraph{Impact of Attention Mechanism}
% Without Attention
% With Attention

With the knowledge retrieved from \textsc{Atomic} or generated from \textsc{Comet}, we are able to infer the possible intentions and reactions of the interlocutors. However, not all knowledge phrases contribute the same to the emotion of the focused utterance. We study the attention mechanism in terms of selecting the relevant knowledge. We show in Table~\ref{tab:4} a heat map of the attention scores in Eq.~\ref{eqs:7} to illustrate how the topic-driven attention could identify the most salient phrase. The utterance `\emph{Oh my God, you're a freak.}' will be erroneously categorized as `\emph{mad}' without using the topic-driven attention (shown in the last row of Table~\ref{tab:4}).  %If we use average pooling, we will still get a `\emph{mad}' label.
In contrast, the attention mechanism guides the model to attend to the more relevant events and thus predict the correct emotion label.
% salient knowledge
%  to aggregate the knowledge phrases
% align with
% As we can see from Table~\ref{tab:4},
% As compared against the pooling method
% It can be observed that
%We draw a heat map in Figure  The results are . Compared with the  The results are printed/presented as heat map from. So that the irrelevant information are discarded . Above: a case study of , below: Attention heatmap of an instance sampled from the testing set. With and without. Compare the classification result
% illustrate
% Error analysis indicates the

\definecolor{reda}{RGB}{255,240,240}
\definecolor{redb}{RGB}{255,235,235}
\definecolor{redc}{RGB}{255,250,250}
\definecolor{redd}{RGB}{255,175,175}
\definecolor{rede}{RGB}{255,210,210}
\definecolor{redf}{RGB}{255,155,155}
\definecolor{redg}{RGB}{255,195,195}
\definecolor{redh}{RGB}{255,225,225}
\definecolor{redi}{RGB}{255,200,200}
\definecolor{redj}{RGB}{255,225,225}
\begin{table}[tp]
\setlength{\fboxsep}{0.5pt}
\DeclareRobustCommand{\hlcyan}[1]{{\sethlcolor{cyan}\hl{#1}}}
\DeclareRobustCommand{\hlreda}[1]{{\sethlcolor{reda}\hl{#1}}}
\DeclareRobustCommand{\hlredb}[1]{{\sethlcolor{redb}\hl{#1}}}
\DeclareRobustCommand{\hlredc}[1]{{\sethlcolor{redc}\hl{#1}}}
\DeclareRobustCommand{\hlredd}[1]{{\sethlcolor{redd}\hl{#1}}}
\DeclareRobustCommand{\hlrede}[1]{{\sethlcolor{rede}\hl{#1}}}
\DeclareRobustCommand{\hlredf}[1]{{\sethlcolor{redf}\hl{#1}}}
\DeclareRobustCommand{\hlredg}[1]{{\sethlcolor{redg}\hl{#1}}}
\DeclareRobustCommand{\hlredh}[1]{{\sethlcolor{redh}\hl{#1}}}
\DeclareRobustCommand{\hlredi}[1]{{\sethlcolor{redi}\hl{#1}}}
\DeclareRobustCommand{\hlredj}[1]{{\sethlcolor{redj}\hl{#1}}}
\begin{center}
\fontsize{9pt}{11pt}\selectfont
\begin{tabular}{c|lc}
\thickhline
\hline
\multicolumn{1}{c|}{\multirow{7}{*}{\parbox[t]{2mm}{\rotatebox[origin=c]{90}{Dialogue Context}}}} &  \multicolumn{1}{c}{\multirow{7}{*}{\tabincell{l}{A: Alright, go on.\\ B: Ok, I have to sleep on the west side\\ \quad\, because I grew up in California\\ \quad\, and otherwise the ocean would be\\ \quad\, on the wrong side.\\ \framebox{{\textcolor{cyan}{A: Oh my God, you're a freak.}}}\\ B: Yeah. How about that.}}} & \multicolumn{1}{c}{\multirow{1}{*}{Neutral}}\\
 &  & \multicolumn{1}{c}{\multirow{1}{*}{Neutral}}\\
 &  & \\
 &  & \\
 &  & \\
 &  & \textcolor{cyan}{Joyful}\\
 &  & Neutral\\
\hline
\multicolumn{1}{c|}{\multirow{9}{*}{\parbox[t]{2mm}{\rotatebox[origin=c]{90}{Topic-Driven Attention}}}} & \hlreda{A wants to be liked} & \multicolumn{1}{c}{\multirow{9}{*}{\tabincell{c}{\textcolor{cyan}{Joyful} \textcolor{red}{\ding{51}}}}}\\
 & \hlredb{A wants to be accepted} & \\
 & \hlredc{A wants to be a freak} & \\
 & \textcolor{black}{\hlredd{A will feel satisfied}} & \\
 & \textcolor{black}{\hlrede{A will feel ashamed}} & \\
 & \textcolor{black}{\hlredf{A will feel happy}} & \\
 & \hlredg{B will feel impressed} & \\
 & \textcolor{black}{\hlredh{B will feel disgusted}} &  \\
 & \textcolor{black}{\hlredi{B will feel surprised}} &  \\
% \hline
% \multicolumn{1}{c|}{\multirow{9}{*}{\parbox[t]{2mm}{\rotatebox[origin=c]{90}{Average Pooling}}}} & \hlredj{A wants to be liked} & \multicolumn{1}{c}{\multirow{9}{*}{\tabincell{c}{\textcolor{forestgreen}{Mad} \textcolor{red}{\ding{55}}}}}\\
%  & \hlredj{A wants to be accepted} & \\
%  & \hlredj{A wants to be a freak} & \\
%  & \textcolor{black}{\hlredj{A will feel satisfied}} & \\
%  & \textcolor{black}{\hlredj{A will feel ashamed}} & \\
%  & \textcolor{black}{\hlredj{A will feel happy}} & \\
%  & \hlredj{B will feel impressed} & \\
%  & \textcolor{black}{\hlredj{B will feel disgusted}} &  \\
%  & \textcolor{black}{\hlredj{B will feel surprised}} &  \\
\hline
 & A: Oh my God, you're a freak. & \textcolor{forestgreen}{Mad} \textcolor{red}{\ding{55}}\\
\thickhline
\end{tabular}
\end{center}
\caption{\label{tab:4} Illustration of the attention mechanism in Eq.~\ref{eqs:7} that helps distinguish the retrieved knowledge.}
\end{table}

\section{Conclusion}
We have proposed a Topic-Driven and Knowledge-Aware Transformer model that incorporates topic representation and the commonsense knowledge from \textsc{Atomic} for emotion detection in dialogues. A topic-augmented language model based on fine-tuning has been developed for topic extraction. Pointer network and additive attention have been explored for knowledge selection. All the novel components have been integrated into the Transformer Encoder-Decoder structure that enables Seq2Seq prediction. Empirical results demonstrate the effectiveness of the model in topic representation learning and knowledge integration, which have both boosted the performance of emotion detection.

\cmrdy{\section*{Acknowledgements}
The authors would like to thank the anonymous reviewers for insightful comments and helpful suggestions. This work was funded by the EPSRC (grant no. EP/T017112/1, EP/V048597/1). LZ is funded by the Chancellor's International Scholarship at the University of Warwick. YH is supported by a Turing AI Fellowship funded by the UK Research and Innovation (grant no. EP/V020579/1). DZ is funded by the National Key Research and Development Program of China (2017YFB1002801) and the National Natural Science Foundation of China (61772132).}

\bibliographystyle{acl_natbib}
\bibliography{anthology,acl2021}

\newpage

\appendix

\section{Appendices}
\subsection{Settings}
We modified the script\footnote{\url{https://huggingface.co/transformers/v2.0.0/examples.html}} of language model fine-tuning \yh{in the} Hugging Face \yh{library}~\cite{wolfetal-2020-transformers} for the implementation of topic-driven fine-tuning. On each training set, we train the topic model for 3 epochs, with learning rate set to $5e$-$5$ to prevent overfitting to the low-resource dataset. The classifier is built on the Transformers\footnote{\url{https://huggingface.co/transformers/}} package in Hugging Face. The language model we employ is RoBERTa~\cite{liu2019roberta}. Each utterance is padded by the \texttt{<pad>} token of RoBERTa if it is less than the maximum length of $128$. The maximum number of utterances in a dialogue is set to $36$, $25$, $72$ and $25$ respectively for DailyDialog~\cite{li2017dailydialog}~\footnote{\url{http://yanran.li/dailydialog.html}}, MELD~\cite{poria2019meld}~\footnote{\url{https://github.com/declare-lab/MELD}}, IEMOCAP~\cite{busso2008iemocap}~\footnote{\url{https://sail.usc.edu/iemocap/iemocap_release.htm}} and EmoryNLP~\cite{zahiri2018emotion}~\footnote{\url{https://github.com/emorynlp/emotion-detection}}. Dialogues with shorter lengths are padded with \texttt{NULL}. It is worth noting that this step is performed after RoBERTa due to the random noises introduced by RoBERTa. The number of \yh{retrieved or generated events from \textsc{Atomic} under the relation types} `\emph{intentions}' and `\emph{reactions}' is both set to $5$, i.e., $K=5$.

%\appendix

\end{document}